%% file: 2023_emnlp_prompt_ensemble.tex
\pdfoutput=1

\documentclass[11pt]{article}

\usepackage{ACL2023}

\usepackage{times}
\usepackage{latexsym}
\usepackage{booktabs}
\usepackage{mdwlist}

\usepackage[T1]{fontenc}

\usepackage[utf8]{inputenc}

\usepackage{microtype}
\usepackage{blindtext}

\usepackage{comment}
\usepackage{inconsolata}
\usepackage{caption}
\usepackage{subcaption}

\newcommand{\abr}[1]{\textsc{#1}}

\title{Getting \underline{MoRE} out of \underline{M}ixture \underline{o}f Language Model \underline{R}easoning \underline{E}xperts
}

\author{Chenglei Si$^{1,4}$ \hspace{1.4cm} 
Weijia Shi$^{2}$ \hspace{1.4cm}  
 \textbf{Chen Zhao}$^{3}$
 \\
 \textbf{Luke Zettlemoyer}$^{2}$
 \hspace{1.0cm}
 \textbf{Jordan Boyd-Graber}$^{1}$ \\
  $^{1}$ University of Maryland \hspace{0.6cm}
  $^{2}$ University of Washington  \\
  \hspace{0.3cm}
  $^{3}$ NYU Shanghai \hspace{0.6cm}
  $^{4}$ Stanford University \\
  \texttt{clsi@stanford.edu}\\
}

\input{style/preamble}
\begin{document}
\maketitle
\input{sections/00-abstract.tex}
\input{sections/10-introduction.tex}
\input{sections/20-background.tex}
\input{sections/30-expert.tex}
\input{sections/40-mope.tex}

\input{sections/50-abstention.tex}
\input{sections/60-related}

\input{sections/70-conclusion.tex}
\input{sections/limitations.tex}
\input{sections/ack.tex}

\bibliography{anthology,custom}
\bibliographystyle{acl_natbib}

\clearpage
\input{sections/appendix.tex}

\end{document}

%% file: style/preamble.tex
\usepackage[a-1b]{pdfx}

\usepackage{framed}
\usepackage{mdwlist}
\usepackage{siunitx}
\usepackage{latexsym}
\usepackage{colortbl}
\usepackage{xcolor}
\usepackage{nicefrac}
\usepackage{booktabs}
\usepackage{fnpct}
\usepackage{amsfonts}
\usepackage[T1]{fontenc}
\usepackage{bold-extra}
\usepackage{amsmath}
\usepackage{amssymb}
\usepackage{bm}
\usepackage{graphicx}
\usepackage{mathtools}
\usepackage{microtype}
\usepackage{multirow}
\usepackage{multicol}
\usepackage{xpatch}
\usepackage{latexsym,comment}
\usepackage[normalem]{ulem}

\newcommand*{\missingreference}{{\Huge \colorbox{red}{?reference?}}}
\newcommand*{\missingcitation}{{\Huge \colorbox{red}{?citation?}}}

\makeatletter
\xpatchcmd{\@setref}{\bfseries}{\missingreference}{}{}
\def\@citex[#1]#2{\leavevmode
    \let\@citea\@empty
    \@cite{\@for\@citeb:=#2\do
        {\@citea\def\@citea{,\penalty\@m\ }%
            \edef\@citeb{\expandafter\@firstofone\@citeb\@empty}%
            \if@filesw\immediate\write\@auxout{\string\citation{\@citeb}}\fi
            \@ifundefined{b@\@citeb}{\hbox{\reset@font\missingcitation}%
                \G@refundefinedtrue
                \@latex@warning
                {Citation `\@citeb' on page \thepage \space undefined}}%
            {\@cite@ofmt{\csname b@\@citeb\endcsname}}}}{#1}}
\makeatother

\newcommand{\gem}[1]{\mbox{\textsc{gem}}}

\newcommand{\hidetext}[1]{}
\newcommand{\ignore}[1]{}

\newcommand{\smallurl}[1]{ \begin{tiny}\url{#1}\end{tiny}}

\definecolor{lightblue}{HTML}{3cc7ea}
\definecolor{CUgold}{HTML}{CFB87C}
\definecolor{grey}{rgb}{0.95,0.95,0.95}
\definecolor{ceil}{rgb}{0.57, 0.63, 0.81}
\definecolor{UMDred}{HTML}{ed1c24}
\definecolor{UMDyellow}{HTML}{ffc20e}

\newcommand{\qa}[0]{\abr{qa}}
\newcommand{\llm}[0]{\abr{llm}}

\newcommand{\nq}[0]{\abr{nq}}

\newcommand{\mope}[0]{\abr{MoRE}}

%% file: sections/00-abstract.tex
\begin{abstract}
While recent large language models (\llm{}s) 
improve on various question answering (\qa{}) datasets, it remains
difficult for a single model to generalize across question types that require
distinct reasoning abilities.
We provide empirical evidence that state-of-the-art \llm{}s 
suffer from poor generalizability on reasoning types beyond
those seen in the prompt.
To remedy this, we propose a Mixture-of-Reasoning-Experts (\mope{}) framework
that ensembles diverse specialized language models. 
We specialize the backbone language model  with prompts optimized
for different reasoning categories,
including factual, multihop, mathematical, and commonsense reasoning. 
Our key insight is to leverage agreement  among the specialized experts to select the best answer for each question, or to abstain from answering. 
This gives \mope{} higher accuracy than any
single specialized model on a collection of 12 \abr{qa} datasets from four reasoning types. Beyond generalizability, the interpretable design of \mope{} improves selective question answering results compared to baselines without incorporating inter-expert agreement.
This framework is also more interpretable and useful to human consumers of \qa{} outputs. Our human study confirms that presenting expert predictions and the
answer selection process helps annotators more accurately calibrate when to
trust the system's output.
We release all code and data to facilitate future work.~\footnote{\href{https://github.com/NoviScl/MoRE}{https://github.com/NoviScl/MoRE}}

\end{abstract}

%% file: sections/10-introduction.tex
\section{Introduction}
\begin{figure}[t]
    \centering
    \includegraphics[scale=0.17]{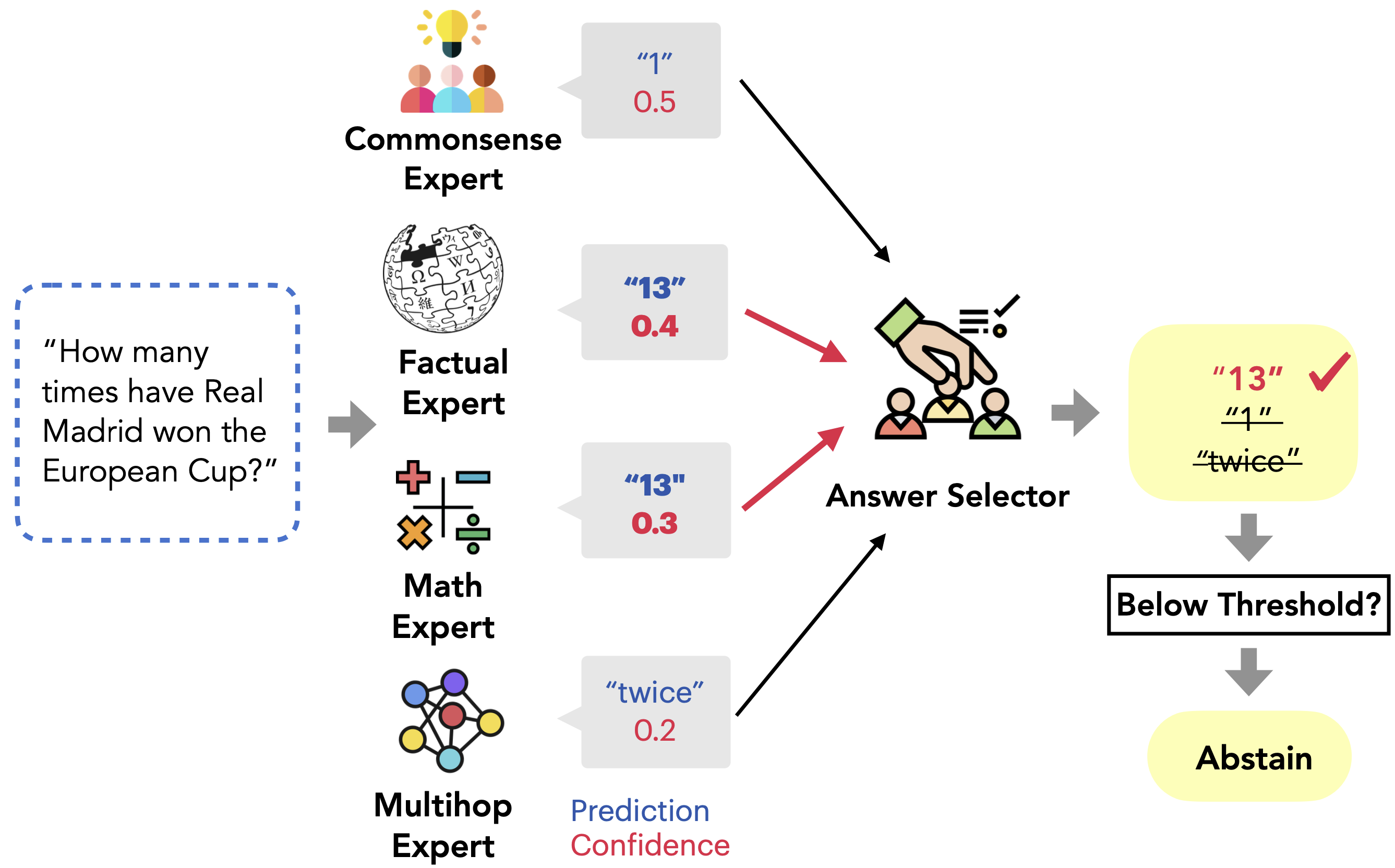}
    \caption{
     \textbf{Overview of \mope{}}. In our \mope{} framework, each of the four specialized expert models produces a prediction for the test question and we train a classifier to select the best answer among them. 
     The answer selector considers all the \emph{predictions} and their \emph{confidence}, as well as their \emph{agreement} (\textit{e.g.}, in this example the factual expert makes the same prediction as the math expert and so this prediction gets a higher score). 
    Finally, if the selected answer's score is relatively low, the system abstains from answering. In this example, the correct answer should be 14, and \mope{} abstained correctly to avoid producing the wrong answer 13. 
    }
    \label{fig:intro}
\end{figure}

Question answering (\qa{}) is one of the most common interactions
between humans
and AI with a wide range of applications~\cite{Gardner2019QuestionAI}.
When a \qa{} system is deployed in-the-wild---where users can ask any
question---the principal challenges are to handle the diversity of
question types while ensuring reliability by only providing answers when the
system has a high probability of being correct.
This motivates us to develop a \abr{qa} system that achieves both goals: (1)
it should be \textbf{generalizable}, adept at handling any type of question;
(2) it should answer \textbf{selectively}, abstaining from producing erroneous answers.

Toward these goals, one popular approach is to build a unified \qa{} system.
While general-purpose LLMs like GPT-3~\cite{OpenAI2021} demonstrate impressive
question-answering abilities, they lack specialization on particular domains or
reasoning types and often fall behind specialized models~\cite{Qin2023IsCA,Kocon2023ChatGPTJO}.
Moreover, to the public, these LLMs are massive black boxes: users have
cannot connect the prediction process to whether the
outputs are trustworthy.

Therefore, we go against this trend of building a single generalist language model,
but rather design a more interpretable system that 
consists of a pool of specialized models and each question is answered by one of them. 
Crucially, to best use complementary strengths of multiple \qa{} models, we 
 implement a pool of diverse and capable specialized models
(\textit{e.g.}, by equipping \llm{}s with corresponding prompting strategies) for
each specific reasoning type; then we train a classifier to select
the best candidate answer from the specialized models for each  question or to abstain from answering
(Figure~\ref{fig:intro}).
This framework, Mixture-of-Reasoning-Experts (\mope{}), aims to both generalize and answer selectively.

To obtain the most capable specialist models for each reasoning type, we
leverage specialized prompting strategies such as
Chain-of-Thought~\cite{Wei2022ChainOT} prompting and retrieval-augmented prompting. 
Experiments on our collection of 12 QA datasets across four diverse reasoning
types confirm that our specialist models outperform the backbone
model without specialization, but they achieve much lower accuracy on question types
outside of their expertise. 


With these specialized models, we propose our \mope{} framework to combine
their strengths.
\mope{} selects the best candidate answer from the pool of
specialized models, and we
teach \mope{} to abstain from answering if none of the candidate answers are correct.
We design our answer selector based on these indicative features: (1) the
match between the question type and each specialized model's expertise; (2)
the confidence of each specialized model and the characteristics of their predictions; and  (3) the agreement among all specialized models, which is a novel feature that we propose.  
Experiments validate that by ensembling the specialized experts this way, \mope{} significantly outperforms any single specialized model
across all four diverse reasoning types.

Apart from the improved generalizability of \mope{}, an important byproduct of cross-checking among specialized experts is to
offer a useful signal for understanding the whole system's working
mechanism.
%
This is validated by the experimental results showing that
incorporating agreement among different specialized experts leads to better
selective \qa{} results---where the system answers as many questions as possible while maintaining high accuracy---and presenting such internal decision processes to
human annotators helps them determine the correctness of the system
predictions more accurately and in a shorter time.






%% file: sections/20-background.tex
\section{Problem Setup}
\label{sec:background}

Given our goal of developing a \qa{} system that generalizes across reasoning types and abstains appropriately, 
we introduce our task and evaluation details. 


\subsection{Generalizablility Across Reasoning Types}

We aim to develop a \abr{qa} system that handles any type of question with different reasoning challenges. Therefore, we evaluate \abr{qa} systems from the following representative reasoning categories:

\begin{itemize*}
    \item \textbf{Factual reasoning}: factoid questions that are knowledge-intensive.

    \item \textbf{Multihop reasoning}: decomposing the question into sub-steps and reasoning across them.

    \item \textbf{Mathematical reasoning}:  mathematical and logical computations, such as math word problems. 

    \item \textbf{Commonsense reasoning}: commonsense knowledge that is often implicit. 
    
\end{itemize*}

Our list of \qa{} reasoning types is selected based on existing \qa{} taxonomy~\cite{Rogers2021QADE}. The list is not exhaustive -- we focus on them partly due to the availability of evaluation benchmarks but our system can be easily extended to other reasoning types.
Our final reported metric is based on the macro-average across 12 different datasets from these reasoning types. 



\subsection{Selective Prediction}

To deploy the \abr{qa} system in real-world applications, the system 
should abstain from answering when its final answer is likely to be wrong.
%
%
Therefore, we adopt the \textbf{\textit{selective}} \qa{} setup~\cite{ElYaniv2010OnTF,Kamath2020SelectiveQA} as our final evaluation setting.\footnote{While our primary focus for evaluation lies in selective \qa{}, in section~\ref{sec:sanity}, we also directly compare predicted answers and gold labels as a sanity check without selective prediction.}
%
More formally, given a question~$x$, the \abr{qa} system returns a predicted answer~$\hat{y}$. 
We assign a score $c \in \mathbb{R}$ to this prediction that reflects the likelihood of this answer being correct. 
We evaluate selective \qa{} by ranking all predictions by their scores $c$ and abstain if the score~$c$ is lower than a threshold~$\gamma$. 
Intuitively, lowering the threshold~$\gamma$ would increase the answering coverage, but also incur higher error rates. We introduce metrics for evaluating such trade-offs in Section~\ref{sec:auto_abstention}.

The crux of the problem is to develop calibrators that can reliably score the predictions to reflect their probability of being correct. This is where the interpretable design of our proposed \mope{} system helps: 
we will demonstrate in Section~\ref{sec:abstention} that the inter-expert agreement information in the \mope{} system is an effective signal for predicting the correctness of answers for both automatic abstention and human verification of answer correctness. 

%% file: sections/30-expert.tex
\section{Mixture of Reasoning Experts}
\label{sec:ensemble}

This section introduces our Mixture of Reasoning Experts (\mope{}) framework, including how to obtain diverse reasoning experts, how to ensemble them,  and how to predict answer correctness. 

\subsection{Specialized Reasoning Experts}

The first step of our \mope{} system is to obtain a diverse set of specialized models so that we can combine their strengths via strategic ensembling. Although there are numerous ways of building specialized \abr{qa} models, we design specialized reasoning experts via prompting a \llm{} since it has state-of-the-art accuracy on many reasoning tasks. 
We specialize the Codex model~\cite{Chen2021EvaluatingLL} for different reasoning types with four specialized prompting methods (the example prompts are listed in the Appendix, Figure~\ref{fig:prompts}):

\begin{itemize*}
 \item \textbf{Factual expert} with retrieval-augmented prompting. Following \citet{Si2022PromptingGT}, for each  question, we retrieve the top 10 most relevant passages from Wikipedia with Contriever~\cite{Izacard2021UnsupervisedDI} and append them to the prompt right before the question. 

 \item \textbf{Multihop expert} with Chain-of-Thought (CoT) prompting~\cite{Wei2022ChainOT}. We add manually-written rationales after each demo question in the prompt to elicit multi-step reasoning process for the questions. 

 \item \textbf{Math expert} with CoT prompting.  We add the accompanied explanations provided in GSM8K after each demo question in the prompt to elicit similar reasoning steps for the questions. 

 \item \textbf{Commonsense expert} with generated knowledge prompting~\cite{Liu2021GeneratedKP}.  We generate 10 fact pieces related to each question using the Codex model and append them to the prompt right before the question. 

\end{itemize*}

After obtaining predictions from each expert, we train a classifier to pick the best answer.
This allows \mope{} to 
ensemble these four specialized expert models 
without knowing \textit{a priori} the question's reasoning type.

\begin{table*}[t]
\tiny
\setlength\tabcolsep{3pt}
\centering
\resizebox{\textwidth}{!}{%
\begin{tabular}{ lccc|ccc|ccc|ccc|c} 
\toprule
&  \multicolumn{3}{c}{Factual} & \multicolumn{3}{c}{Multihop} & \multicolumn{3}{c}{Math} & \multicolumn{3}{c}{Commonsense} & \\
\midrule
& NQ & TQA & SQuAD & HQA & BeerQA3+ & MuSiQue & GSM8K & SVAMP & MultiArith & CSQA & CSQA2.0 & QASC & Macro-Average \\
\midrule
\textit{Single Expert Results (Section~\ref{sec:loss_generalizability})} \\
Specific Few-Shot & 37.8 & 70.3 & 20.0 & 27.3 & 31.5 & 10.3 & 19.5 & 66.0 & 41.5 & 75.8 & 64.0 & 67.4  & 44.3 \\
Factual Expert & \textbf{42.8} & \textbf{72.3} & \textbf{30.0} & \textbf{37.0} & 27.0 & 12.5 & 11.8 & 53.5 & 32.2 & 46.6 & 62.0 & 33.1 & 38.4 \\
Multihop Expert &  34.8 & 61.3 & 19.0 & 34.3 & \textbf{46.3} & \textbf{15.5} & 37.5 & 70.5 & 75.9 & 55.2 & 62.5 & 54.1 & 47.2 \\
Math Expert & 21.0 & 59.8 & 13.8 & 22.5 & 34.0 & 7.5 & \textbf{61.8} & \textbf{74.5} & \textbf{92.2} & 51.1 & 58.0 & 57.9 & 46.2 \\
Commonsense Expert & 32.5 & 64.0 & 16.3 & 31.3 & 38.5 & 10.8 & 41.5 & 72.5 & 75.4 & \textbf{78.4} & \textbf{65.3} & \textbf{68.9} & 49.6 \\
\midrule
\textit{Ensemble: Full \mope{} Router (Section~\ref{sec:mope_gene})} \\
Oracle & 53.8 & 78.5 & 37.0 & 51.7 & 61.0 & 25.5 & 75.8 & 90.3 & 99.2 & 92.1 & 86.0 & 88.2 & \underline{69.9} \\
Majority Vote & 33.8 & 68.0 & 18.3 & 31.3 & 33.0 & 9.0 & 26.5 & 64.5 & 68.8 & 57.0 & 63.0 & 49.6 & 43.6 \\
MaxProb & 38.8 & 69.3 & 23.3 & 38.5 & 42.5 & 13.5 & 48.5 & 75.3 & 83.9 & 47.6 & 62.0 & 53.4 & 49.7 \\
\mope{} - Codex Router & 34.5 & 62.7 & 18.5 & 36.0 & 45.3 & 15.3 & 53.8 & 77.0 & 88.7 & 60.8 & 63.0 & 60.7 & 51.4 \\
\mope{} - RF Router & 39.0 & 71.8 & 25.8 & 37.5 & 46.0 & 14.0 & 63.5 & 80.5 & 95.0 & 78.9 & 66.8 & 72.9 & \textbf{57.6} \\
\midrule
\textit{Ensemble: Question Only Router (Section~\ref{sec:question_only})} \\
Random Selector Baseline & 32.3 & 64.8 & 21.5 & 33.3 & 37.3 & 10.5 & 35.8 & 67.8 & 70.6 & 54.2 & 62.0 & 53.6 & 45.3 \\
Q-Type Oracle & 42.8 & 72.3 & 30.0 & 34.3 & 46.3 & 15.5 & 61.8 & 74.5 & 92.2 & 78.4 & 65.3 & 68.9 & 56.8 \\
\mope{} - RF Router & 34.5 & 62.7 & 20.5 & 31.5 & 39.0 & 10.8 & 52.3 & 74.3 & 89.2 & 67.7 & 63.5 & 56.4 & 50.2 \\
 \bottomrule
\end{tabular}}
\caption{
Per-dataset accuracy (exact match) breakdown on all 12 \qa{} datasets. 
We highlight the best single-expert result on each dataset in \textbf{bold}.
Specialized \qa{} models (first block) excel at the corresponding reasoning types and lose generalizability on others. 
Our proposed \mope{} system with the random forest answer selector (second block) has the best macro-average accuracy across all datasets (57.6), beating all specialized \qa{} models, although it still lags behind the oracle ensemble (69.9).  \mope{} with the few-shot Codex router 
 performs significantly worse than the full random forest router (51.4). 
 Notably, \mope{} with the question-only random forest router (last block) can still outperform the single expert baselines but performs much worse than the full \mope{} router.
}
\label{tab:per_dataset}
\end{table*}

\subsection{Ensembling via Answer Selection}
\label{sec:router}

We combine the strengths of the specialized experts by employing a feature-based random forest classifier to score each candidate answer, the score is used for selecting the final answer and determining when to abstain.
We assume the setting where we obtain the predictions from each specialized model first and then select the best answer. 
We describe the details of training the classifier in this section. 

\paragraph{Feature Set}

We use hand-designed features including the expert type, question characteristics (\textit{e.g.}, the question word, length, and existence of numerical values), answer characteristics (\textit{e.g.}, confidence, length, and the token overlap with questions, contexts, and rationales), and inter-expert agreement. We include the full list of features in the Appendix (Section~\ref{sec:features}). Here we highlight the inter-expert agreement 
features 
that are uniquely introduced in this work thanks to the more interpretable design of \mope{}, which includes the frequency of the predicted answer among all four experts' predictions, and the token overlap among these expert predictions. 

Additionally, we experiment with a setting where we route the  question to the best expert based only on the question itself without obtaining predictions from all experts. In that setting, we train the random forest classifier without using any answer characteristic or inter-expert answer agreement features  (more details in Section~\ref{sec:question_only}).

\paragraph{Training Data and Objective}  
We hold out 100 examples per \qa{} dataset as the training data (1200 examples in total). 
During training, we extract the features from the questions and the expert models' outputs to train the random forest classifier with a binary classification objective to predict whether the expert model prediction is correct or not. 
During inference, for each question, we score all experts' answers with this classifier and select the answer with the highest score as the final answer. If the final selected answer's score is below a searched threshold, we abstain from answering.  

Apart from the random forest classifier, we also experimented with other feature-based classifiers and finetuning pretrained language models like BERT~\cite{Devlin2019BERTPO}, but found them to be less effective. 

\paragraph{Few-Shot Answer Selection}  
While training a random forest answer selector gives better \qa{} accuracy (as we will show in the next section), it requires a moderate amount of training data. We also explore a few-shot alternative, where we directly prompt the Codex model with 14 randomly selected demo examples,
each consisting of the question, the predictions of the four specialized models, and the best answer among them.~\footnote{We only select examples where the correct answer is among the expert predictions.} During inference, we append the question and prompt Codex to select the best answer.

%% file: sections/40-mope.tex
\section{Sanity Check: \mope{} Improves Generalizability}
\label{sec:sanity}

This section describes our experiments to verify that \mope{}'s ensemble of diverse experts improves generalizability. 

\label{sec:mope}

\subsection{Experimental Setup}

\paragraph{Evaluation Datasets}
We evaluate on 12 datasets covering four reasoning types. Specifically, Natural Questions (NQ)~\cite{Kwiatkowski2019NaturalQA}, TriviaQA~\cite{Joshi2017TriviaQAAL}, and SQuAD~\cite{Rajpurkar2016SQuAD1Q} for factual reasoning; 
HotpotQA~\cite{Yang2018HotpotQAAD}, BeerQA~\cite{Qi2020AnsweringOQ},\footnote{The original BeerQA dataset contains a mixture of single-hop and multi-hop questions, we only take the 3+ hops subset and name it BeerQA3+ for our evaluation. } and MuSiQue~\cite{Trivedi2021MM} for multihop reasoning;
GSM8K~\cite{Cobbe2021TrainingVT}, SVAMP~\cite{Patel2021AreNM}, and MultiArith~\cite{Roy2016SolvingGA} for mathematical reasoning; 
CommonsenseQA (CSQA)~\cite{Talmor2019CommonsenseQAAQ}, CSQA2.0~\cite{Talmor2021CommonsenseQA2E}, and QASC~\cite{Khot2019QASCAD} for commonsense reasoning. 
For each dataset, we randomly sample 400 questions from the test set for our evaluation to control inference costs. 

\paragraph{Demonstration Examples for Specialized Experts}
We use 16 randomly sampled training examples as demonstration examples of each specialized prompt. 
Specifically, we use examples from Natural Questions as demonstration examples for the factual expert, examples from 
HotpotQA for the multihop expert, examples GSM8K for the mach expert, and examples from CSQA for the commonsense expert. These demonstration examples are formatted with the corresponding specialized prompting strategies described above. 
Additionally, we also include a dataset-specific few-shot baseline where we randomly sample and concatenate 16 question-answer pairs from each corresponding dataset being evaluated as the prompt without any specialized prompting techniques. 
We use the answer exact match (EM) as the evaluation metric for all datasets.



\subsection{Specialization and Loss of Generalizability}
\label{sec:loss_generalizability}

We first evaluate each of the four specialized reasoning experts on the collection of 12 datasets (Table~\ref{tab:per_dataset}, first block). 

\paragraph{The specialized experts excel at their targeted reasoning types.} For example, the factual expert outperforms the dataset-specific few-shot baseline on NQ, TriviaQA, and SQuAD, and the math expert improves accuracy from 19.5 to 61.8 on GSM8K and from 41.5 to 92.2 on MultiArith. The only exception is that the factual expert is the best-performing model on HotpotQA---a multihop reasoning benchmark. This is because HotpotQA is also knowledge-intensive~\cite{Yang2018HotpotQAAD}, and retrieval augmentation can be even more helpful than  Chain-of-Thought reasoning. 

\paragraph{The specialized experts are worse on reasoning types outside of their expertise.} For instance, the factual expert underperforms the dataset-specific few-shot baseline on all math and commonsense datasets. Similarly, the math expert underperforms the baseline on all factoid \qa{} datasets. This means that a single specialized \qa{} model cannot generalize on the diverse types of questions and it motivates us to propose the \mope{} system to combine the strengths of different experts in order to fare well on all types of reasoning questions.

\subsection{\mope{} Improves Generalizability}
\label{sec:mope_gene}

Here we focus on the full \mope{} router that scores each expert's answer for answer selection.
The second block in Table~\ref{tab:per_dataset} compares \mope{} with several other baselines:

\begin{itemize*}
    \item \textbf{Oracle Ensemble}: We compute the upper bound by taking the optimal answer for each question. Therefore, for each question, as long as one of the expert models got the correct answer, the accuracy will be 1. 

    \item \textbf{Majority Vote}: We choose the most frequent answer string among the four expert models as the final prediction. 

    \item \textbf{MaxProb}: We choose the answer with the highest confidence score.
\end{itemize*}

\mope{} with either the Codex answer selector or the random forest selector has better macro-average accuracy on the 12 datasets than any of the single-expert baselines (the first block in Table~\ref{tab:per_dataset}) and is also better than the majority vote or MaxProb baseline. 
In particular, \mope{} with the random forest selector 
beats the best-performing expert (Commonsense Expert) by 8 points in macro-average accuracy and is significantly better than the Codex selector, demonstrating strong generalizability. 

We emphasize that {\bf we do not know the question type beforehand}.
The single expert baselines do excel at their corresponding question types (e.g., factual expert performs the best on factual questions, even better than MoRE), but they perform terribly on other question types (e.g., the factual expert is much worse than normal dataset-specific few-shot prompting on math and commonsense questions).
In contrast, for any given test question, our MoRE system’s answer selector can select the best expert for that question without prior knowledge of its type. This selection process is crucial because there is no single expert model excelling across all types of questions. Therefore, it is this generalization accuracy (i.e., the “macro-average” accuracy column in Table~\ref{tab:per_dataset}) that we are highlighting as MoRE’s core advantage, where MoRE scores 57.6 accuracy, outperforming all single-expert baselines in macro-average accuracy by large margins.

\subsection{Question-Only Routing}
\label{sec:question_only}

In this section, we introduce the \textbf{Question-Only} setting, where we route based on the question alone. This means that we do not ask all four expert models for an answer; instead, we pick one expert and get the answer from it.
Thus, we train the random forest router without any features that involve the expert predictions or their agreement.
We also include two baselines for this setting: 1) randomly selecting an expert for each question; 2) a question-type oracle where we always route the question to the expert specialized in the corresponding question type (\textit{e.g.}, we route all factual questions to the factual expert and all multi-hop questions to the multi-hop expert; which assumes knowledge of the question types).  

This setup contrasts with the full \mope{} router setting in Section~\ref{sec:mope_gene}, where all four expert models answer and then select the best answer.
This requires four times more compute, but allows us to obtain more information for the expert selection.

The question-only routing approach beats single-expert baselines (Table~\ref{tab:per_dataset}, last block), but lags behind full \mope{}.
In particular, \mope{}'s question-only router has a macro-average accuracy of 50.2, slightly higher than the best single-expert (Commonsense Expert) with 49.6 accuracy, but significantly lower than \mope{} with the full router (57.6 macro-average accuracy). In the remaining sections of the paper, we focus only on the \mope{} router given its strong performance, and study how to enable selective prediction.

%% file: sections/50-abstention.tex
\section{\mope{} Improves Selective QA}
\label{sec:abstention}

The previous section has confirmed the generalizability strength of \mope{}, but it is still far from perfect. In fact, it is impossible for any \qa{} system to be perfectly accurate on all questions, thus highlighting the importance of abstention---the system should not output an answer when it is likely to be wrong. 
For this goal, \mope{} has the important advantage of being more interpretable since users can understand how the system derives the final answer by inspecting each expert's prediction and the answer selection process. We demonstrate the benefits of such interpretability via evaluation on automatic abstention as well as human abstention.





\subsection{Automatic Abstention}
\label{sec:auto_abstention}

\begin{table*}[t]
\setlength\tabcolsep{6pt}
\centering
\begin{tabular}{ lcccc } 
 \toprule
Method &  
 AUC$_\downarrow$ & Cov@Acc=80\%$_\uparrow$ & Cov@Acc=90\%$_\uparrow$ & ER$_\uparrow$
 \\
  \midrule
  MaxProb & 34.8 & 32.4 & 12.4 & 17.5 \\
  RF Calibrator w/o Agreement & 36.0 & 26.6 & 12.8 & 22.9 \\
  \mope{} Calibrator & 28.3 & 45.9 & 34.3 & 33.4 \\
 \bottomrule
\end{tabular}
\caption{
  Incorporating inter-expert agreement features in the \mope{} calibrator improves selective \qa{} as measured by all metrics and outperforms the MaxProb baseline by large margins. All results are the macro-average over 12 datasets.
}
\label{tab:calibration}
\end{table*}



\begin{table*}[t]
\small
\setlength\tabcolsep{5pt}
\centering
\begin{tabular}{ l | cc | cc | cc | c } 
 \toprule
Condition &  Decision Acc & ER  & Accept Correct & Reject Wrong & Correct Conf & Wrong Conf & Time (Mins/20Qs) \\
  \midrule
Baseline & 57.0 & 9.5 & 75.0 & 36.8 & 0.69 & 0.59 & 15.0 \\
\mope{} & 67.5 & 19.5 & 89.4 & 43.8 & 0.78 & 0.67  & 13.2 \\
 \bottomrule
\end{tabular}
\caption{
 In human studies, 20 annotators (200 annotations) decide whether the system prediction is correct: 1) They achieve higher accuracy in deciding whether the final system output is correct when presented with information about the expert predictions and their scores (the \mope{} condition), which also corresponds to higher effective reliability (ER). 2) Showing expert information improves annotators' accuracy in both accepting correct answers and rejecting wrong predictions; 3) It boosts user confidence in both their correct and wrong judgments (although ideally we want the confidence on wrong judgments to be lower); 4) The \mope{} condition also takes less time for users to make decisions. 
}
\label{tab:human_calibration}
\end{table*}

\begin{figure*}[t]
\centering
\includegraphics[width=1.0\textwidth]{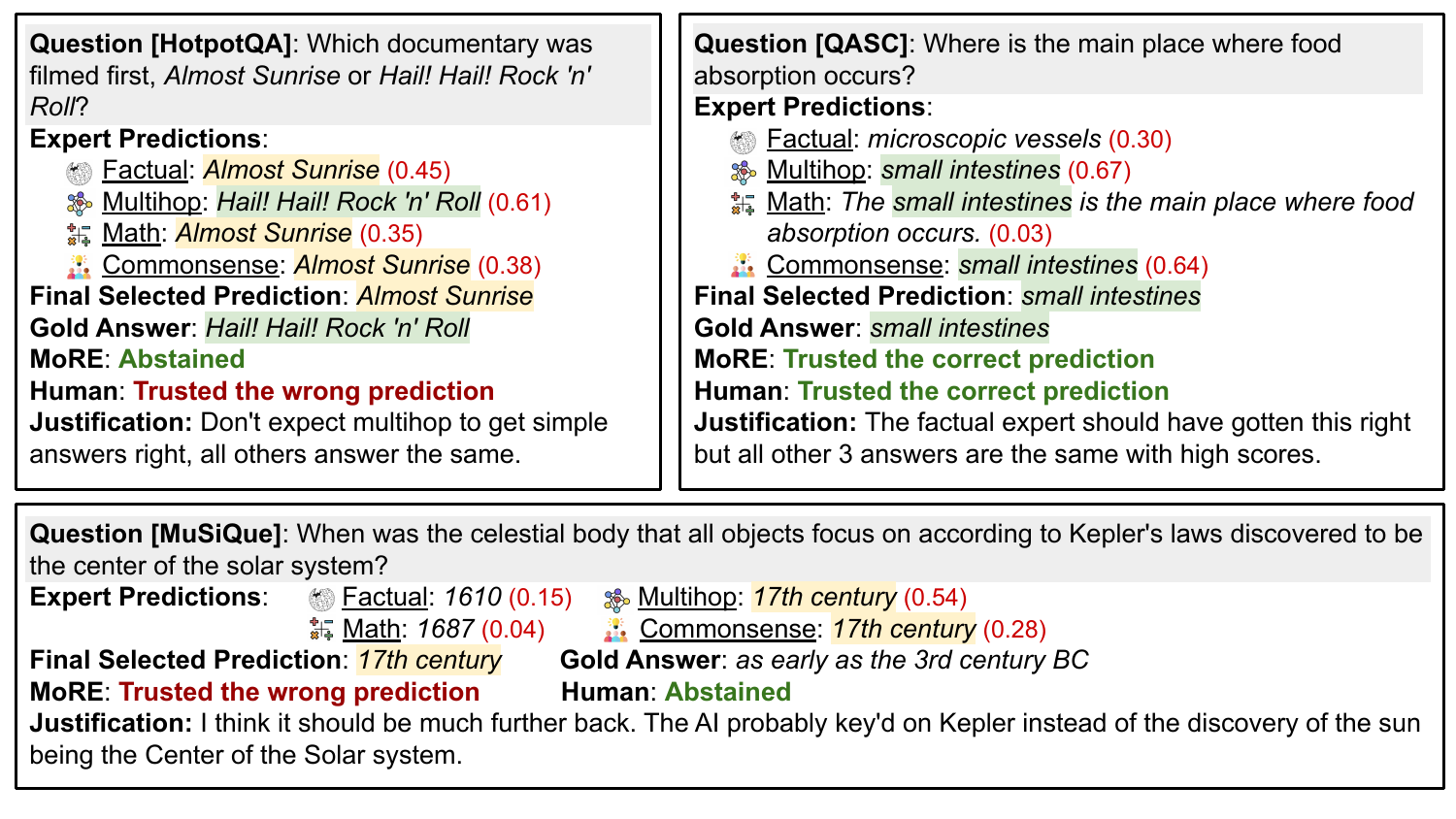}
\caption{ 
Three examples of \mope{} automatic abstention and human abstention. For each example, we show the question, each reasoning expert's prediction along with its score, the best prediction selected by the random forest classifier, the actual gold answer, the abstention decision by \mope{} and human annotators as well as the annotators' justification.   
\textbf{Humans often rely on inter-expert agreement and  their own understanding of how these expert models work.} 
}
\label{fig:examples}
\end{figure*}





Traditionally, the decision to abstain or not is determined based solely on a confidence score. 
However, confidence scores of the generated answers can be poorly calibrated~\cite{Jiang2020HowCW,Si2022RevisitingCF} for this purpose. 
A more effective approach is to train a calibrator to score the prediction's probability of being correct~\cite{Kamath2020SelectiveQA,Ye2021CanEB,zhang2021knowing}. 
For \mope{}, we can easily use the answer selector as the calibrator to score the final predictions. 
Since \mope{} gathers predictions from multiple experts, it enables users to take advantage of the agreement among these expert systems as an additional useful signal apart from the confidence scores. 
To verify the effectiveness of such inter-expert agreement signals, we use the random forest selector from Section~\ref{sec:router} to score model predictions, and ablate the impact of including inter-expert agreement features. 
We use the same \mope{} system with the random forest selector as the underlying \qa{} system, which means that the \qa{} accuracy would stay the same across all settings. 
We then compare the following three ways of scoring the final system predictions for automatic abstention: 

\begin{itemize*}
    \item \textbf{MaxProb}: We directly take the selected answer's language modeling probability (as provided by the underlying Codex model) as the prediction's score. 

    \item \textbf{RF Calibrator w/o Inter-Expert Agreement}: To tease apart the impact of inter-expert agreement features, we train the random forest classifier without any of the inter-expert agreement features described in Section~\ref{sec:router}. 

    \item \textbf{\mope{} Calibrator}: We use the random forest classifier with all features in Section~\ref{sec:router} as the calibrator. We simply take the classifier's predicted score on the selected answer as the score for the final prediction.  
\end{itemize*}

We use the following established metrics for evaluating the effectiveness of selective \qa{}: 

\begin{itemize*}
    \item \textbf{Area Under Curve (AUC)}: For any given threshold $\gamma$, there is an associated coverage and error rate (risk).  We plot risk versus coverage and
    evaluate the area under this curve (AUC). This metric averages over
    all possible threshold $\gamma$, and lower AUC indicates better selective \abr{qa} performance. 

    \item \textbf{Coverage at Accuracy (Cov@Acc)}: We report the maximum possible coverage for a desired accuracy level. We report Cov@80\% and Cov@90\% in the table. 

    \item \textbf{Effective Reliability (ER)}: Following~\citet{Whitehead2022ReliableVQ}, we compute the score $\phi$ of each prediction as: (1) $\phi = 1$ if the system chooses to output an answer and the answer is correct (exact match equals 1); (2) $\phi = 0$ if the system chooses to abstain; (3) $\phi = -1$ is the system chooses to output an answer but the answer is wrong. The ER is then computed as the average of this score over the test set of size $n$: $\Phi = \frac{1}{n} \sum_x \phi(x)$. The threshold $\gamma$ for deciding whether to abstain or not is tuned on our dev set (which consists of 100 questions from each dataset) and applied on the test sets. 
\end{itemize*}




\paragraph{Results}

Our full \mope{} calibrator wins on on all metrics (Table~\ref{tab:calibration}) including AUC, Cov@80\%, Cov@90\%, and effective reliability. Interestingly, the random forest calibrator without the inter-expert agreement features is worse than the MaxProb baseline (e.g., on AUC), which further highlights the benefit of having the inter-expert agreement as part of the calibrator design.

\subsection{Human Abstention}

We next verify that the expert-agreement and answer-selection information also help humans determine the correctness of the system's output. 

\paragraph{Setup}
For the human study, 
we recruit 20 annotators from Prolific, who each annotated 20 randomly sampled questions.
Our  between-subject study has two conditions: (1) in the \textbf{baseline} condition, we present users with only the question and the final \mope{} prediction; (2) in the \textbf{\mope{}} condition, apart from the question and the final answer, 
 we also present the predictions of each expert model along with the random forest classifier's scores of the candidate answers (interface in Appendix Figure~\ref{fig:interface}). 
 We also include a brief description of every expert's specialization in the task instruction to help annotators better understand the information. 
 Half of the annotators were assigned to the baseline condition and the other half to the \mope{} condition. 
 We provide an average compensation of \$14.7 per hour and  did not apply any additional screening apart from asking for proficient English speakers.

\paragraph{Results}
For each question, we ask annotators to decide: (1) whether they think the final prediction is correct (binary judgment); and (2) what is their confidence in their own judgment on a scale of 1 to 5, which we will convert to a numerical value in range $[0, 1]$ for computing the average. 

\mope{} improves both the accuracy and efficiency of human answer verification (Table~\ref{tab:human_calibration}).
\mope{} improves annotators' accuracy of deciding whether the system prediction is correct from 57.0\% to 67.5\% ($p=0.012$), which also corresponds to a jump in effective reliability from 9.5 to 19.5. When we break down the results into the accuracy of accepting correct model predictions and rejecting wrong model predictions, the \mope{} condition improves accuracy in both categories. When measuring annotators' confidence in their judgment, their confidence increases in both correct and wrong judgment, as a result of seeing the additional inter-expert agreement information in the \mope{} condition. 

Lastly and somewhat surprisingly, \mope{}'s additional information did not slow people down: annotators spend an average of 13.2 minutes every 20 questions, compared to the average time of 15.0 minutes in the baseline condition, possibly because the lack of supporting evidence in the baseline condition makes the decision process difficult for people (similar effect as in~\citet{Feng2022LearningTE}).   
Interestingly, the automatic calibrator from \mope{} has an effective reliability score of 11.3 on the same sampled set annotators saw. This is higher than the human ER in the baseline condition (9.5) but lower than the human ER in the \mope{} condition, indicating that humans are able to capture additional cues that our automatic calibrator missed. Next, we examine those cases where humans effectively overruled \mope{}.

\paragraph{Case Studies}
While humans largely rely on the expert selection and inter-expert agreement for abstention like the \mope{} calibrator, they sometimes also use background knowledge about the question (examples in  Figure~\ref{fig:examples}). In the first example from HotpotQA, the annotator trusted the wrong prediction because three of the expert models made the same prediction and the annotator didn't recognize that the question is multihop (in fact the multihop gave the correct answer but it's not selected as the final prediction). In the second example from QASC, although the annotator judged the question to be a factoid question, they went with the consensus of the other three expert models. These two examples show that humans rely on both the match between the question type and corresponding expert strength, as well as the inter-expert agreement for their judgment. In the third example from MuSiQue, the annotator inferred why the model made the particular prediction and successfully spotted the mistake. Such external knowledge may partially account for why humans get better abstention effective reliability than \mope{}. 

%% file: sections/60-related.tex
\section{Related Work}



\paragraph{Specialized Prompting and Prompt Ensemble} To better elicit knowledge and reasoning from LLMs, many prompting methods have been proposed, such as Least-to-Most~\cite{Zhou2022LeasttoMostPE} and Self-Ask Prompting~\cite{Press2022MeasuringAN} for multi-step reasoning, and Program-of-Thought~\cite{Chen2022ProgramOT} and Declarative Prompting~\cite{Ye2023SatisfiabilityAidedLM} for symbolic reasoning. 
Unlike these works, our goal is to combine the strengths of all the specialized language models empowered with these specialized prompting techniques for better generalizability and selective QA. 
Another line of work ensembles multiple answers from LLMs: \citet{Wang2022SelfConsistencyIC} samples multiple answers with a high temperature during decoding and selects the final answer by majority vote; while \citet{Li2022MakingLL} constructs different prompts by selecting different demonstration examples and trains a verifier to perform weighted voting on the answers. Unlike these approaches, we create reasoning experts with different specializations in order to achieve generalizability and leverage the inter-expert agreement features for both answer selection and abstention. 

\paragraph{Modular LM and Mixture-of-Experts}
One classic example towards modular language models is Mixture-of-Experts~\cite{Jacobs1991AdaptiveMO}, which is adopted in scaling sparse Transformer models like GShard~\cite{Lepikhin2020GShardSG}, Switch-Transformer~\cite{Fedus2021SwitchTS},  BASELayer~\cite{Lewis2021BASELS}, DEMIX~\cite{Gururangan2021DEMixLD}, Branch-Train-Merge~\cite{Li2022BranchTrainMergeEP}, and C-BTM~\cite{Gururangan2023ScalingEL}.
Unlike these Mixture-of-Experts, our \mope{} system does not route at the token level but rather designs specialized experts and routes the entire question to the best expert. The most similar works to ours are~\citet{Puerto2021MetaQACE} and \citet{Jiang2023LLMBlenderEL}, where each expert model generates an entire response to the query and a reranker then selects the best answer.
However, unlike all these prior works, each specialized model in \mope{} is carefully designed to excel in a particular reasoning type (rather than domain experts like most prior works), allowing for better complementary strengths across reasoning types, and to the best of our knowledge, we are the first study to focus on ensembling experts under the more practical selective \abr{qa} setting.

\paragraph{Generalizable QA and Multitask Learning}  MRQA~\cite{Fisch2019MRQA2S} benchmarked the domain generalizability of machine reading comprehension models and similar to \citet{Talmor2019MultiQAAE}: \qa{} models trained on one domain often fail to generalize on others. To improve generalizability, \citet{Khashabi2020UnifiedQACF} trained a unified model on a large collection of \qa{} datasets, while \citet{Friedman2021SingledatasetEF} trained lightweight adapters for domain generalization. 
Unlike these works, we focus on the more challenging setting of generalizing across different reasoning types, and we take a different approach by ensembling multiple specialized models. 
Beyond \qa{}, a growing line of work trains multitask models via multitask training~\cite{Zhong2021AdaptingLM,Min2021MetaICLLT} or instruction tuning~\cite{Mishra2021CrossTaskGV,Wei2021FinetunedLM,Wang2022SuperNaturalInstructionsGV}, which allows \llm{}s to extrapolate across different types of tasks. 
However, such fine-tuned models (with multitask or instruction tuning) still suffer from poor interpretability, while our proposed framework allows users to inspect the internal expert selection process for better interpretability.

\paragraph{Selective Prediction}
Several prior works studied training effective calibrators to decide when to abstain. 
\citet{Kamath2020SelectiveQA} studied selective \qa{} under domain shifts where they showed that training a random forest calibrator is better than relying on LM probability alone. \citet{Ye2021CanEB} additionally included local explanation features to improve the calibrator, and \citet{zhang2021knowing} embedded questions as dense vector features to improve the calibrator. \citet{Xie2022CalibratingTO} focused specifically on multihop questions and achieved benefits from incorporating question decomposition information in the calibrator. 
\citet{Garg2021WillTQ} filtered unanswerable questions based on model confidence to improve computation efficiency. 
\citet{Rodriguez2019QuizbowlTC} studied incremental question answering (Quizbowl) where calibration is an intrinsic part of the task in order to decide the best timing for making a prediction (``buzzing''). 
Our work contributes to this line of work by showing the benefit of designing a more interpretable \qa{} system where the inter-expert agreement features are helpful for calibration and selective \qa{}.

%% file: sections/70-conclusion.tex
\section{Conclusion}

We proposed the \mope{} framework where we construct a pool of specialized \qa{} models that excel at different reasoning types, and then train an answer selector to select the best answer among them. Experiments on 12 datasets covering four reasoning types demonstrate that \mope{} achieve better generalizability than all baselines. More importantly, the inter-expert agreement features in \mope{} offer useful signals for training effective calibrators that improve selective \qa{} and also improve human verification of the system's final predictions. 

While we focused on prompting \llm{}s as specialized experts, the idea of combining the strengths of diverse experts can extend to any type of specialized models, even non-neural ones such as traditional information retrieval models, which is an interesting avenue for future work. 
Additionally, future work could also explore other possible explanations to facilitate users' calibration and abstention, such as better explaining the strengths and weaknesses of individual specialized expert models. Such efforts are especially important for high-stakes settings that require careful fact-checking or verification of the system outputs~\cite{HumanVerification}. 

\section*{Limitations}
\paragraph{Model Coverage} We only focused on the Codex model for the experiments due to its strong performance on \qa{} tasks (at the time of writing this paper). It would be interesting to verify our framework on different \llm{}s, especially open-source models.  Moreover, future work could move beyond using prompted \llm{}s as the specialized experts and instead ensemble more heterogeneous expert models such as models finetuned on particular reasoning types or non-Transformer models.

\paragraph{Reasoning Type Coverage} We experimented with four representative reasoning types but there exist many more question types that could possibly occur in real-life applications, such as questions with multiple answers, ambiguous questions, and questions with false presuppositions. It would be interesting for future work to study how to extend our framework to also tackle these additional reasoning types, for example by designing and adding new specialized models.

\paragraph{Beyond \qa{}} While we only focused on \qa{} evaluation, another interesting direction for future work is to extend our idea beyond just \qa{}, for example for general-purpose language modeling. This likely requires re-designing the evaluation pipeline and implementing specialized expert models that are not only performant for \qa{} tasks but for language generation in general. 

\section*{Ethical Considerations}

\paragraph{Human Study} Our human study has been exempted by the Institutional Review Boards, and we compensate annotators an average of \$14.7 per hour, well above the minimum wage in the US. We do not expect any harm during the entire annotation process. 

\paragraph{Broader Impact} Our work improves the reliability of \qa{} systems in the wild and improves the long-standing problem of users over-trusting answers from black-box AI systems. 
We believe that our interpretable \mope{} system can inspire more future work on designing AI systems where humans can verify the answers and calibrate their trust appropriately in order to avoid being misled by erroneous AI outputs. 

%% file: sections/limitations.tex


%% file: sections/ack.tex
\section*{Acknowledgement}
We thank Ruiqi Zhong, Tianyu Gao, Xi Ye, and Daniel Khashabi for their helpful discussion. 
We thank Peng Qi for providing the BeerQA evaluation data. 
We thank Navita Goyal for providing helpful advice on building the human study interface. 
Chenglei Si completed this work back when he was an undergraduate researcher at UMD and he thanks all members of the UMD CLIP lab for their support throughout his undergraduate research journey. 
Chen Zhao is supported by Shanghai Frontiers Science Center of Artificial Intelligence and Deep Learning, NYU Shanghai. 
This work is also supported by Meta AI through Dynabench Data Collection and Benchmarking Platform.

%% file: sections/appendix.tex
\appendix
\section{Appendix}

\subsection{Prompts for Reasoning Experts}

\begin{figure*}
\centering
\includegraphics[width=1.0\textwidth]{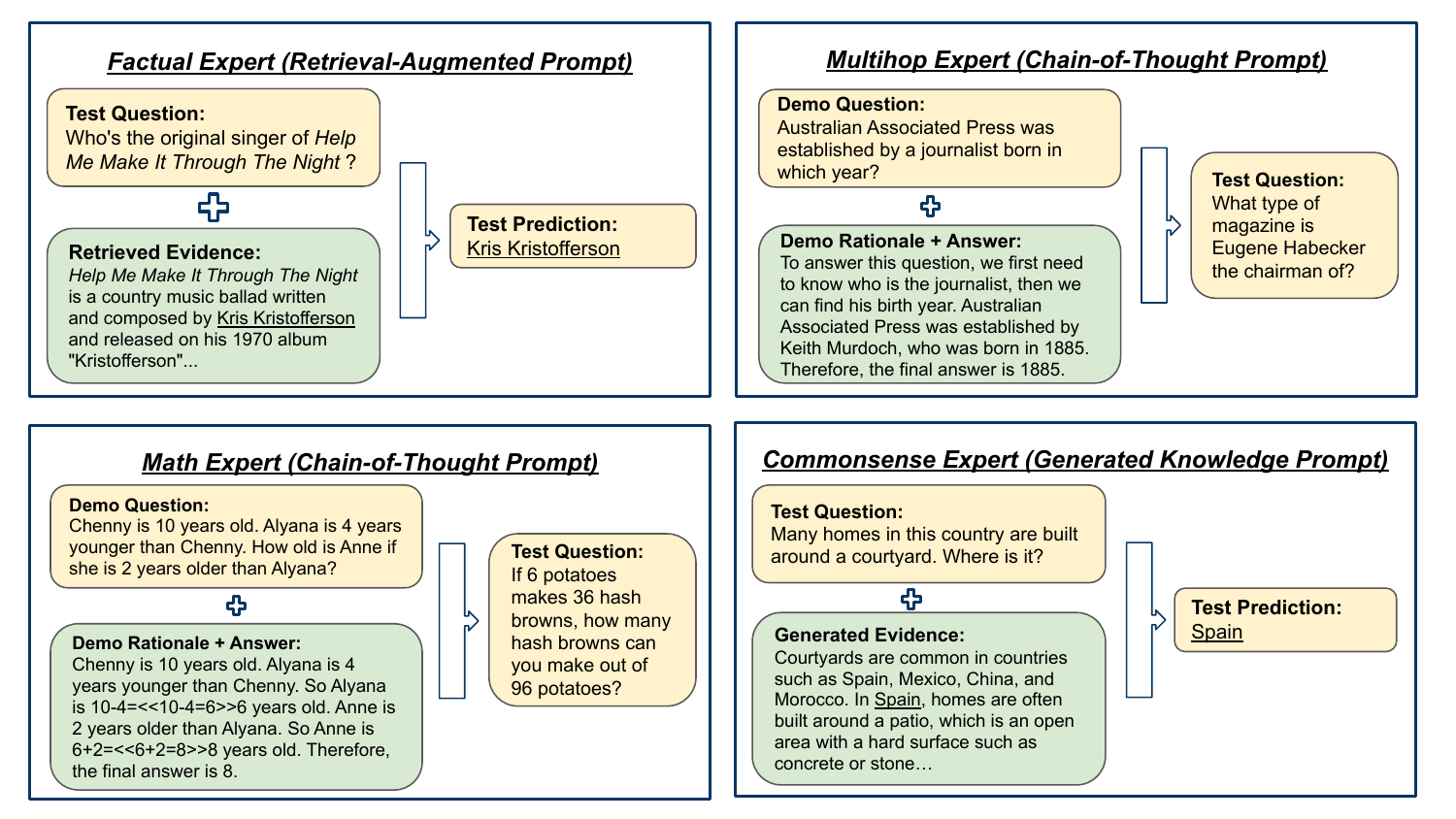}
\caption{The four specialized \abr{qa} models in \mope{}, implemented by applying specialized prompts on Codex. For the factual expert, the demo examples are randomly sampled examples from \nq{} and we append retrieved evidence from Wikipedia for each test question; for the multihop expert, we use question and rationale-answer pairs from HotpotQA as the prompt; for the math expert, we use question and rationale-answer pairs from GSM8K as the prompt; for commonsense expert, we use random examples from CommonsenseQA as the prompt, and we use the same LLM to generate related background knowledge to append to each test question.  
}
\label{fig:prompts}
\end{figure*}

Figure~\ref{fig:prompts} shows the actual prompt design for the four specialized reasoning experts in \mope{}.

\subsection{Interface for Human Study}

\begin{figure*}
\centering
\includegraphics[trim={5cm 13cm 4cm 5cm}, width=0.9\textwidth]{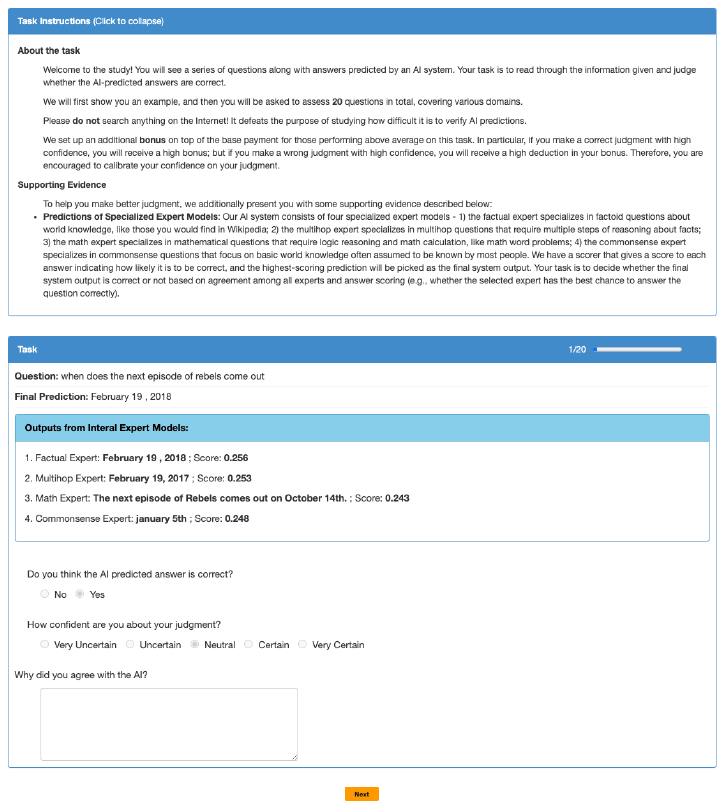}
\caption{
Our annotation interface for the human abstention study. 
}
\label{fig:interface}
\end{figure*}

Figure~\ref{fig:interface} shows the annotation interface for our human abstention study. We provide instructions for the task, describe the reasoning experts' strengths, then show the test questions with all expert predictions and scores. We then ask annotators to determine the correctness of the final prediction, their confidence, as well as their justification. In the baseline condition, the expert prediction panel is omitted.

\subsection{Features for Training the Classifier}
\label{sec:features}

Below we list all the features used to train our random forest classifier that scores expert predictions.

\begin{itemize*}
    \item \textbf{Specialized Expert Type}: a one-hot four-dimensional vector. 

    \item \textbf{Question Characteristics}: question word, question length, and the number of numerical values in the question.

    \item \textbf{Answer Characteristics}: the probability of the generated output
    (multiplying each token's likelihood and normalizing by length as in~\citet{Si2022PromptingGT}), 
    the length of the generated answer, the overlap between the question and the predicted answer, the number of numerical values in the answer, overlap between the answer and retrieved or generated passages, length of CoT rationales, 
    overlap between questions and rationales,
    overlap between answers and rationales, 
    the number of times the answer appears in the rationale, 
    and the number of numerical values in the rationale.

    \item \textbf{Factual and Commonsense Experts' Contexts}: the number of numerical values in the retrieved or generated passages, the number of overlapping tokens between questions and passages, and the passage length. 

    \item \textbf{Inter-Expert Agreement}: the frequency of the predicted answer among all four experts' predictions, token overlap among the experts' outputs. 
    
\end{itemize*}

Some of these features are expanded upon prior works on selected \qa{}~\cite{Rodriguez2019QuizbowlTC,Ye2021CanEB,zhang2021knowing}.

\clearpage

%% file: 2023_emnlp_prompt_ensemble.bbl
\begin{thebibliography}{59}
\expandafter\ifx\csname natexlab\endcsname\relax\def\natexlab#1{#1}\fi

\bibitem[{Chen et~al.(2021)Chen, Tworek, Jun, Yuan, Ponde, Kaplan, Edwards,
  Burda, Joseph, Brockman, Ray, Puri, Krueger, Petrov, Khlaaf, Sastry, Mishkin,
  Chan, Gray, Ryder, Pavlov, Power, Kaiser, Bavarian, Winter, Tillet, Such,
  Cummings, Plappert, Chantzis, Barnes, Herbert-Voss, Guss, Nichol, Babuschkin,
  Balaji, Jain, Carr, Leike, Achiam, Misra, Morikawa, Radford, Knight,
  Brundage, Murati, Mayer, Welinder, McGrew, Amodei, McCandlish, Sutskever, and
  Zaremba}]{Chen2021EvaluatingLL}
Mark Chen, Jerry Tworek, Heewoo Jun, Qiming Yuan, Henrique Ponde, Jared Kaplan,
  Harrison Edwards, Yura Burda, Nicholas Joseph, Greg Brockman, Alex Ray, Raul
  Puri, Gretchen Krueger, Michael Petrov, Heidy Khlaaf, Girish Sastry, Pamela
  Mishkin, Brooke Chan, Scott Gray, Nick Ryder, Mikhail Pavlov, Alethea Power,
  Lukasz Kaiser, Mohammad Bavarian, Clemens Winter, Philippe Tillet,
  Felipe~Petroski Such, David~W. Cummings, Matthias Plappert, Fotios Chantzis,
  Elizabeth Barnes, Ariel Herbert-Voss, William~H. Guss, Alex Nichol, Igor
  Babuschkin, S.~Arun Balaji, Shantanu Jain, Andrew Carr, Jan Leike, Joshua
  Achiam, Vedant Misra, Evan Morikawa, Alec Radford, Matthew~M. Knight, Miles
  Brundage, Mira Murati, Katie Mayer, Peter Welinder, Bob McGrew, Dario Amodei,
  Sam McCandlish, Ilya Sutskever, and Wojciech Zaremba. 2021.
\newblock {Evaluating Large Language Models Trained on Code}.
\newblock \emph{arXiv}, abs/2107.03374.

\bibitem[{Chen et~al.(2022)Chen, Ma, Wang, and Cohen}]{Chen2022ProgramOT}
Wenhu Chen, Xueguang Ma, Xinyi Wang, and William~W. Cohen. 2022.
\newblock {Program of Thoughts Prompting: Disentangling Computation from
  Reasoning for Numerical Reasoning Tasks}.
\newblock \emph{arXiv}, abs/2211.12588.

\bibitem[{Cobbe et~al.(2021)Cobbe, Kosaraju, Bavarian, Hilton, Nakano, Hesse,
  and Schulman}]{Cobbe2021TrainingVT}
Karl Cobbe, Vineet Kosaraju, Mohammad Bavarian, Jacob Hilton, Reiichiro Nakano,
  Christopher Hesse, and John Schulman. 2021.
\newblock {Training Verifiers to Solve Math Word Problems}.
\newblock \emph{arXiv}, abs/2110.14168.

\bibitem[{Devlin et~al.(2019)Devlin, Chang, Lee, and
  Toutanova}]{Devlin2019BERTPO}
Jacob Devlin, Ming-Wei Chang, Kenton Lee, and Kristina Toutanova. 2019.
\newblock {BERT: Pre-training of Deep Bidirectional Transformers for Language
  Understanding}.
\newblock In \emph{Proceedings of NAACL}.

\bibitem[{El-Yaniv and Wiener(2010)}]{ElYaniv2010OnTF}
Ran El-Yaniv and Yair Wiener. 2010.
\newblock {On the Foundations of Noise-free Selective Classification}.
\newblock \emph{Journal of Machine Learning Research}, 11:1605--1641.

\bibitem[{Fedus et~al.(2022)Fedus, Zoph, and Shazeer}]{Fedus2021SwitchTS}
William Fedus, Barret Zoph, and Noam~M. Shazeer. 2022.
\newblock {Switch Transformers: Scaling to Trillion Parameter Models with
  Simple and Efficient Sparsity}.
\newblock \emph{Journal of Machine Learning Research}.

\bibitem[{Feng and Boyd-Graber(2022)}]{Feng2022LearningTE}
Shi Feng and Jordan Boyd-Graber. 2022.
\newblock {Learning to Explain Selectively: A Case Study on Question
  Answering}.
\newblock In \emph{Conference on Empirical Methods in Natural Language
  Processing}.

\bibitem[{Fisch et~al.(2019)Fisch, Talmor, Jia, Seo, Choi, and
  Chen}]{Fisch2019MRQA2S}
Adam Fisch, Alon Talmor, Robin Jia, Minjoon Seo, Eunsol Choi, and Danqi Chen.
  2019.
\newblock {MRQA 2019 Shared Task: Evaluating Generalization in Reading
  Comprehension}.
\newblock In \emph{Proceedings of EMNLP}.

\bibitem[{Friedman et~al.(2021)Friedman, Dodge, and
  Chen}]{Friedman2021SingledatasetEF}
Dan Friedman, Ben Dodge, and Danqi Chen. 2021.
\newblock {Single-dataset Experts for Multi-dataset Question Answering}.
\newblock In \emph{Proceedings of EMNLP}.

\bibitem[{Gardner et~al.(2019)Gardner, Berant, Hajishirzi, Talmor, and
  Min}]{Gardner2019QuestionAI}
Matt Gardner, Jonathan Berant, Hannaneh Hajishirzi, Alon Talmor, and Sewon Min.
  2019.
\newblock {Question Answering is a Format; When is it Useful?}
\newblock \emph{arXiv}, abs/1909.11291.

\bibitem[{Garg and Moschitti(2021)}]{Garg2021WillTQ}
Siddhant Garg and Alessandro Moschitti. 2021.
\newblock {Will this Question be Answered? Question Filtering via Answer Model
  Distillation for Efficient Question Answering}.
\newblock In \emph{Conference on Empirical Methods in Natural Language
  Processing}.

\bibitem[{Gururangan et~al.(2021)Gururangan, Lewis, Holtzman, Smith, and
  Zettlemoyer}]{Gururangan2021DEMixLD}
Suchin Gururangan, Michael Lewis, Ari Holtzman, Noah~A. Smith, and Luke
  Zettlemoyer. 2021.
\newblock {DEMix Layers: Disentangling Domains for Modular Language Modeling}.
\newblock In \emph{Proceedings of NAACL}.

\bibitem[{Gururangan et~al.(2023)Gururangan, Li, Lewis, Shi, Althoff, Smith,
  and Zettlemoyer}]{Gururangan2023ScalingEL}
Suchin Gururangan, Margaret Li, Mike Lewis, Weijia Shi, Tim Althoff, Noah~A.
  Smith, and Luke Zettlemoyer. 2023.
\newblock {Scaling Expert Language Models with Unsupervised Domain Discovery}.
\newblock \emph{arXiv}, abs/2303.14177.

\bibitem[{Izacard et~al.(2022)Izacard, Caron, Hosseini, Riedel, Bojanowski,
  Joulin, and Grave}]{Izacard2021UnsupervisedDI}
Gautier Izacard, Mathilde Caron, Lucas Hosseini, Sebastian Riedel, Piotr
  Bojanowski, Armand Joulin, and Edouard Grave. 2022.
\newblock {Unsupervised Dense Information Retrieval with Contrastive Learning}.
\newblock \emph{Transactions on Machine Learning Research}.

\bibitem[{Jacobs et~al.(1991)Jacobs, Jordan, Nowlan, and
  Hinton}]{Jacobs1991AdaptiveMO}
Robert~A. Jacobs, Michael~I. Jordan, Steven~J. Nowlan, and Geoffrey~E. Hinton.
  1991.
\newblock {Adaptive Mixtures of Local Experts}.
\newblock \emph{Neural Computation}, 3:79--87.

\bibitem[{Jiang et~al.(2023)Jiang, Ren, and Lin}]{Jiang2023LLMBlenderEL}
Dongfu Jiang, Xiang Ren, and Bill~Yuchen Lin. 2023.
\newblock {LLM-Blender: Ensembling Large Language Models with Pairwise Ranking
  and Generative Fusion}.
\newblock In \emph{Proceedings of ACL}.

\bibitem[{Jiang et~al.(2020)Jiang, Araki, Ding, and Neubig}]{Jiang2020HowCW}
Zhengbao Jiang, J.~Araki, Haibo Ding, and Graham Neubig. 2020.
\newblock {How Can We Know When Language Models Know? On the Calibration of
  Language Models for Question Answering}.
\newblock \emph{Transactions of the Association for Computational Linguistics},
  9:962--977.

\bibitem[{Joshi et~al.(2017)Joshi, Choi, Weld, and
  Zettlemoyer}]{Joshi2017TriviaQAAL}
Mandar Joshi, Eunsol Choi, Daniel~S. Weld, and Luke Zettlemoyer. 2017.
\newblock {TriviaQA: A Large Scale Distantly Supervised Challenge Dataset for
  Reading Comprehension}.
\newblock In \emph{Proceedings of ACL}.

\bibitem[{Kamath et~al.(2020)Kamath, Jia, and Liang}]{Kamath2020SelectiveQA}
Amita Kamath, Robin Jia, and Percy Liang. 2020.
\newblock {Selective Question Answering under Domain Shift}.
\newblock In \emph{Proceedings of ACL}.

\bibitem[{Khashabi et~al.(2020)Khashabi, Min, Khot, Sabharwal, Tafjord, Clark,
  and Hajishirzi}]{Khashabi2020UnifiedQACF}
Daniel Khashabi, Sewon Min, Tushar Khot, Ashish Sabharwal, Oyvind Tafjord,
  Peter Clark, and Hannaneh Hajishirzi. 2020.
\newblock {UnifiedQA: Crossing Format Boundaries With a Single QA System}.
\newblock In \emph{Findings of EMNLP}.

\bibitem[{Khot et~al.(2019)Khot, Clark, Guerquin, Jansen, and
  Sabharwal}]{Khot2019QASCAD}
Tushar Khot, Peter Clark, Michal Guerquin, Peter~Alexander Jansen, and Ashish
  Sabharwal. 2019.
\newblock {QASC: A Dataset for Question Answering via Sentence Composition}.
\newblock In \emph{Proceedings of AAAI}.

\bibitem[{Koco'n et~al.(2023)Koco'n, Cichecki, Kaszyca, Kochanek, Szydlo,
  Baran, Bielaniewicz, Gruza, Janz, Kanclerz, Koco'n, Koptyra,
  Mieleszczenko-Kowszewicz, Milkowski, Oleksy, Piasecki, Radli'nski, Wojtasik,
  Wo'zniak, and Kazienko}]{Kocon2023ChatGPTJO}
Jan Koco'n, Igor Cichecki, Oliwier Kaszyca, Mateusz Kochanek, Dominika Szydlo,
  Joanna Baran, Julita Bielaniewicz, Marcin Gruza, Arkadiusz Janz, Kamil
  Kanclerz, Anna Koco'n, Bartlomiej Koptyra, Wiktoria Mieleszczenko-Kowszewicz,
  P.~Milkowski, Marcin Oleksy, Maciej Piasecki, Lukasz Radli'nski, Konrad
  Wojtasik, Stanislaw Wo'zniak, and Przemyslaw Kazienko. 2023.
\newblock {ChatGPT: Jack of all trades, master of none}.
\newblock \emph{arXiv}, abs/2302.10724.

\bibitem[{Kwiatkowski et~al.(2019)Kwiatkowski, Palomaki, Redfield, Collins,
  Parikh, Alberti, Epstein, Polosukhin, Devlin, Lee, Toutanova, Jones, Kelcey,
  Chang, Dai, Uszkoreit, Le, and Petrov}]{Kwiatkowski2019NaturalQA}
Tom Kwiatkowski, Jennimaria Palomaki, Olivia Redfield, Michael Collins,
  Ankur~P. Parikh, Chris Alberti, Danielle Epstein, Illia Polosukhin, Jacob
  Devlin, Kenton Lee, Kristina Toutanova, Llion Jones, Matthew Kelcey, Ming-Wei
  Chang, Andrew~M. Dai, Jakob Uszkoreit, Quoc~V. Le, and Slav Petrov. 2019.
\newblock Natural questions: A benchmark for question answering research.
\newblock \emph{Transactions of the Association for Computational Linguistics},
  7:453--466.

\bibitem[{Lepikhin et~al.(2020)Lepikhin, Lee, Xu, Chen, Firat, Huang, Krikun,
  Shazeer, and Chen}]{Lepikhin2020GShardSG}
Dmitry Lepikhin, HyoukJoong Lee, Yuanzhong Xu, Dehao Chen, Orhan Firat, Yanping
  Huang, Maxim Krikun, Noam Shazeer, and Zhifeng Chen. 2020.
\newblock {GShard: Scaling Giant Models with Conditional Computation and
  Automatic Sharding}.
\newblock In \emph{Proceedings of ICLR}.

\bibitem[{Lewis et~al.(2021)Lewis, Bhosale, Dettmers, Goyal, and
  Zettlemoyer}]{Lewis2021BASELS}
Mike Lewis, Shruti Bhosale, Tim Dettmers, Naman Goyal, and Luke Zettlemoyer.
  2021.
\newblock {BASE Layers: Simplifying Training of Large, Sparse Models}.
\newblock In \emph{ICML}.

\bibitem[{Li et~al.(2022{\natexlab{a}})Li, Gururangan, Dettmers, Lewis,
  Althoff, Smith, and Zettlemoyer}]{Li2022BranchTrainMergeEP}
Margaret Li, Suchin Gururangan, Tim Dettmers, Mike Lewis, Tim Althoff, Noah~A.
  Smith, and Luke Zettlemoyer. 2022{\natexlab{a}}.
\newblock {Branch-Train-Merge: Embarrassingly Parallel Training of Expert
  Language Models}.
\newblock \emph{ArXiv}, abs/2208.03306.

\bibitem[{Li et~al.(2022{\natexlab{b}})Li, Lin, Zhang, Fu, Chen, Lou, and
  Chen}]{Li2022MakingLL}
Yifei Li, Zeqi Lin, Shizhuo Zhang, Qiang Fu, B.~Chen, Jian-Guang Lou, and
  Weizhu Chen. 2022{\natexlab{b}}.
\newblock {Making Large Language Models Better Reasoners with Step-Aware
  Verifier}.
\newblock \emph{arXiv}, abs/2206.02336.

\bibitem[{Liu et~al.(2021)Liu, Liu, Lu, Welleck, West, Bras, Choi, and
  Hajishirzi}]{Liu2021GeneratedKP}
Jiacheng Liu, Alisa Liu, Ximing Lu, Sean Welleck, Peter West, Ronan~Le Bras,
  Yejin Choi, and Hannaneh Hajishirzi. 2021.
\newblock {Generated Knowledge Prompting for Commonsense Reasoning}.
\newblock In \emph{Proceedings of ACL}.

\bibitem[{Min et~al.(2022)Min, Lewis, Zettlemoyer, and
  Hajishirzi}]{Min2021MetaICLLT}
Sewon Min, Mike Lewis, Luke Zettlemoyer, and Hannaneh Hajishirzi. 2022.
\newblock {MetaICL: Learning to Learn In Context}.
\newblock In \emph{Proceedings of NAACL}.

\bibitem[{Mishra et~al.(2021)Mishra, Khashabi, Baral, and
  Hajishirzi}]{Mishra2021CrossTaskGV}
Swaroop Mishra, Daniel Khashabi, Chitta Baral, and Hannaneh Hajishirzi. 2021.
\newblock {Cross-Task Generalization via Natural Language Crowdsourcing
  Instructions}.
\newblock In \emph{Proceedings of ACL}.

\bibitem[{OpenAI(2022)}]{OpenAI2021}
OpenAI. 2022.
\newblock {Introducing ChatGPT}.
\newblock \url{https://openai.com/blog/chatgpt}.

\bibitem[{Patel et~al.(2021)Patel, Bhattamishra, and Goyal}]{Patel2021AreNM}
Arkil Patel, S.~Bhattamishra, and Navin Goyal. 2021.
\newblock {Are NLP Models really able to Solve Simple Math Word Problems?}
\newblock In \emph{Proceedings of NAACL}.

\bibitem[{Press et~al.(2022)Press, Zhang, Min, Schmidt, Smith, and
  Lewis}]{Press2022MeasuringAN}
Ofir Press, Muru Zhang, Sewon Min, Ludwig Schmidt, Noah~A. Smith, and Mike
  Lewis. 2022.
\newblock Measuring and narrowing the compositionality gap in language models.
\newblock \emph{arXiv}, abs/2210.03350.

\bibitem[{Puerto et~al.(2023)Puerto, Sahin, and Gurevych}]{Puerto2021MetaQACE}
Haritz Puerto, G{\"o}zde~G{\"u}l Sahin, and Iryna Gurevych. 2023.
\newblock {MetaQA: Combining Expert Agents for Multi-Skill Question Answering}.
\newblock In \emph{Proceedings of EACL}.

\bibitem[{Qi et~al.(2020)Qi, Lee, Sido, and Manning}]{Qi2020AnsweringOQ}
Peng Qi, Haejun Lee, Oghenetegiri~TG Sido, and Christopher~D. Manning. 2020.
\newblock {Answering Open-Domain Questions of Varying Reasoning Steps from
  Text}.
\newblock In \emph{Proceedings of EMNLP}.

\bibitem[{Qin et~al.(2023)Qin, Zhang, Zhang, Chen, Yasunaga, and
  Yang}]{Qin2023IsCA}
Chengwei Qin, Aston Zhang, Zhuosheng Zhang, Jiaao Chen, Michihiro Yasunaga, and
  Diyi Yang. 2023.
\newblock {Is ChatGPT a General-Purpose Natural Language Processing Task
  Solver?}
\newblock In \emph{Proceedings of EMNLP}.

\bibitem[{Rajpurkar et~al.(2016)Rajpurkar, Zhang, Lopyrev, and
  Liang}]{Rajpurkar2016SQuAD1Q}
Pranav Rajpurkar, Jian Zhang, Konstantin Lopyrev, and Percy Liang. 2016.
\newblock {SQuAD: 100,000+ Questions for Machine Comprehension of Text}.
\newblock In \emph{Proceedings of EMNLP}.

\bibitem[{Rodriguez et~al.(2019)Rodriguez, Feng, Iyyer, He, and
  Boyd-Graber}]{Rodriguez2019QuizbowlTC}
Pedro Rodriguez, Shi Feng, Mohit Iyyer, He~He, and Jordan~L. Boyd-Graber. 2019.
\newblock Quizbowl: The case for incremental question answering.
\newblock \emph{Journal of Machine Learning Research}.

\bibitem[{Rogers et~al.(2021)Rogers, Gardner, and Augenstein}]{Rogers2021QADE}
Anna Rogers, Matt Gardner, and Isabelle Augenstein. 2021.
\newblock Qa dataset explosion: A taxonomy of nlp resources for question
  answering and reading comprehension.
\newblock \emph{ACM Computing Surveys}, 55:1 -- 45.

\bibitem[{Roy and Roth(2015)}]{Roy2016SolvingGA}
Subhro Roy and Dan Roth. 2015.
\newblock {Solving general arithmetic word problems}.
\newblock In \emph{Proceedings of EMNLP}.

\bibitem[{Si et~al.(2023{\natexlab{a}})Si, Gan, Yang, Wang, Wang, Boyd-Graber,
  and Wang}]{Si2022PromptingGT}
Chenglei Si, Zhe Gan, Zhengyuan Yang, Shuohang Wang, Jianfeng Wang, Jordan
  Boyd-Graber, and Lijuan Wang. 2023{\natexlab{a}}.
\newblock {Prompting GPT-3 To Be Reliable}.
\newblock In \emph{Proceedings of ICLR}.

\bibitem[{Si et~al.(2023{\natexlab{b}})Si, Navita~Goyal, Zhao, Feng, III, and
  Boyd-Graber}]{HumanVerification}
Chenglei Si, Sherry Tongshuang~Wu Navita~Goyal, Chen Zhao, Shi Feng, Hal~Daumé
  III, and Jordan Boyd-Graber. 2023{\natexlab{b}}.
\newblock {Large Language Models Help Humans Verify Truthfulness -- Except When
  They Are Convincingly Wrong}.
\newblock \emph{ArXiv}, abs/2310.12558.

\bibitem[{Si et~al.(2022)Si, Zhao, Min, and Boyd-Graber}]{Si2022RevisitingCF}
Chenglei Si, Chen Zhao, Sewon Min, and Jordan Boyd-Graber. 2022.
\newblock {Revisiting Calibration for Question Answering}.
\newblock \emph{Findings of EMNLP}.

\bibitem[{Talmor and Berant(2019)}]{Talmor2019MultiQAAE}
Alon Talmor and Jonathan Berant. 2019.
\newblock {MultiQA: An Empirical Investigation of Generalization and Transfer
  in Reading Comprehension}.
\newblock In \emph{Proceedings of ACL}.

\bibitem[{Talmor et~al.(2019)Talmor, Herzig, Lourie, and
  Berant}]{Talmor2019CommonsenseQAAQ}
Alon Talmor, Jonathan Herzig, Nicholas Lourie, and Jonathan Berant. 2019.
\newblock {CommonsenseQA: A Question Answering Challenge Targeting Commonsense
  Knowledge}.
\newblock In \emph{Proceedings of NAACL}.

\bibitem[{Talmor et~al.(2021)Talmor, Yoran, Le~Bras, Bhagavatula, Goldberg,
  Choi, and Berant}]{Talmor2021CommonsenseQA2E}
Alon Talmor, Ori Yoran, Ronan Le~Bras, Chandra Bhagavatula, Yoav Goldberg,
  Yejin Choi, and Jonathan Berant. 2021.
\newblock {CommonsenseQA 2.0: Exposing the Limits of AI through Gamification}.
\newblock In \emph{Proceedings of NeurIPS}.

\bibitem[{Trivedi et~al.(2021)Trivedi, Balasubramanian, Khot, and
  Sabharwal}]{Trivedi2021MM}
H.~Trivedi, Niranjan Balasubramanian, Tushar Khot, and Ashish Sabharwal. 2021.
\newblock {MuSiQue: Multihop Questions via Single-hop Question Composition}.
\newblock \emph{Transactions of the Association for Computational Linguistics},
  10:539--554.

\bibitem[{Wang et~al.(2023)Wang, Wei, Schuurmans, Le, hsin Chi, and
  Zhou}]{Wang2022SelfConsistencyIC}
Xuezhi Wang, Jason Wei, Dale Schuurmans, Quoc Le, Ed~Huai hsin Chi, and Denny
  Zhou. 2023.
\newblock {Self-Consistency Improves Chain of Thought Reasoning in Language
  Models}.
\newblock In \emph{Proceedings of ICLR}.

\bibitem[{Wang et~al.(2022)Wang, Mishra, Alipoormolabashi, Kordi, Mirzaei,
  Arunkumar, Ashok, Dhanasekaran, Naik, Stap, Pathak, Karamanolakis, Lai,
  Purohit, Mondal, Anderson, Kuznia, Doshi, Patel, Pal, Moradshahi, Parmar,
  Purohit, Varshney, Kaza, Verma, Puri, Karia, Sampat, Doshi, Mishra, Reddy,
  Patro, Dixit, Shen, Baral, Choi, Smith, Hajishirzi, and
  Khashabi}]{Wang2022SuperNaturalInstructionsGV}
Yizhong Wang, Swaroop Mishra, Pegah Alipoormolabashi, Yeganeh Kordi, Amirreza
  Mirzaei, Anjana Arunkumar, Arjun Ashok, Arut~Selvan Dhanasekaran, Atharva
  Naik, David Stap, Eshaan Pathak, Giannis Karamanolakis, Haizhi~Gary Lai,
  Ishan Purohit, Ishani Mondal, Jacob Anderson, Kirby Kuznia, Krima Doshi,
  Maitreya Patel, Kuntal~Kumar Pal, M.~Moradshahi, Mihir Parmar, Mirali
  Purohit, Neeraj Varshney, Phani~Rohitha Kaza, Pulkit Verma, Ravsehaj~Singh
  Puri, Rushang Karia, Shailaja~Keyur Sampat, Savan Doshi, Siddharth~Deepak
  Mishra, Sujan Reddy, Sumanta Patro, Tanay Dixit, Xudong Shen, Chitta Baral,
  Yejin Choi, Noah~A. Smith, Hanna Hajishirzi, and Daniel Khashabi. 2022.
\newblock {Super-NaturalInstructions: Generalization via Declarative
  Instructions on 1600+ NLP Tasks}.
\newblock In \emph{Conference on Empirical Methods in Natural Language
  Processing}.

\bibitem[{Wei et~al.(2022{\natexlab{a}})Wei, Bosma, Zhao, Guu, Yu, Lester, Du,
  Dai, and Le}]{Wei2021FinetunedLM}
Jason Wei, Maarten Bosma, Vincent Zhao, Kelvin Guu, Adams~Wei Yu, Brian Lester,
  Nan Du, Andrew~M. Dai, and Quoc~V. Le. 2022{\natexlab{a}}.
\newblock {Finetuned Language Models Are Zero-Shot Learners}.
\newblock In \emph{Proceedings of ICLR}.

\bibitem[{Wei et~al.(2022{\natexlab{b}})Wei, Wang, Schuurmans, Bosma, Xia, Chi,
  Le, Zhou et~al.}]{Wei2022ChainOT}
Jason Wei, Xuezhi Wang, Dale Schuurmans, Maarten Bosma, Fei Xia, Ed~H Chi,
  Quoc~V Le, Denny Zhou, et~al. 2022{\natexlab{b}}.
\newblock {Chain-of-Thought Prompting Elicits Reasoning in Large Language
  Models}.
\newblock In \emph{Proceedings of NeurIPS}.

\bibitem[{Whitehead et~al.(2022)Whitehead, Petryk, Shakib, Gonzalez, Darrell,
  Rohrbach, and Rohrbach}]{Whitehead2022ReliableVQ}
Spencer Whitehead, Suzanne Petryk, Vedaad Shakib, Joseph~E. Gonzalez, Trevor
  Darrell, Anna Rohrbach, and Marcus Rohrbach. 2022.
\newblock Reliable visual question answering: Abstain rather than answer
  incorrectly.
\newblock In \emph{Proceedings of ECCV}.

\bibitem[{Xie et~al.(2022)Xie, Wiegreffe, and Riedl}]{Xie2022CalibratingTO}
Kaige Xie, Sarah Wiegreffe, and Mark~O. Riedl. 2022.
\newblock {Calibrating Trust of Multi-Hop Question Answering Systems with
  Decompositional Probes}.
\newblock In \emph{Proceedings of EMNLP}.

\bibitem[{Yang et~al.(2018)Yang, Qi, Zhang, Bengio, Cohen, Salakhutdinov, and
  Manning}]{Yang2018HotpotQAAD}
Zhilin Yang, Peng Qi, Saizheng Zhang, Yoshua Bengio, William~W. Cohen, Ruslan
  Salakhutdinov, and Christopher~D. Manning. 2018.
\newblock {HotpotQA: A Dataset for Diverse, Explainable Multi-hop Question
  Answering}.
\newblock In \emph{Proceedings of EMNLP}.

\bibitem[{Ye et~al.(2023)Ye, Chen, Dillig, and
  Durrett}]{Ye2023SatisfiabilityAidedLM}
Xi~Ye, Qiaochu Chen, Işıl Dillig, and Greg Durrett. 2023.
\newblock {Satisfiability-Aided Language Models Using Declarative Prompting}.
\newblock In \emph{Proceedings of NeurIPS}.

\bibitem[{Ye and Durrett(2021)}]{Ye2021CanEB}
Xi~Ye and Greg Durrett. 2021.
\newblock {Can Explanations Be Useful for Calibrating Black Box Models?}
\newblock In \emph{Proceedings of ACL}.

\bibitem[{Zhang et~al.(2021)Zhang, Gong, and Choi}]{zhang2021knowing}
Shujian Zhang, Chengyue Gong, and Eunsol Choi. 2021.
\newblock {Knowing More About Questions Can Help: Improving Calibration in
  Question Answering}.
\newblock In \emph{Findings of ACL}.

\bibitem[{Zhong et~al.(2021)Zhong, Lee, Zhang, and Klein}]{Zhong2021AdaptingLM}
Ruiqi Zhong, Kristy Lee, Zheng Zhang, and Dan Klein. 2021.
\newblock {Adapting Language Models for Zero-shot Learning by Meta-tuning on
  Dataset and Prompt Collections}.
\newblock In \emph{Conference on Empirical Methods in Natural Language
  Processing}.

\bibitem[{Zhou et~al.(2023)Zhou, Scharli, Hou, Wei, Scales, Wang, Schuurmans,
  Bousquet, Le, and hsin Chi}]{Zhou2022LeasttoMostPE}
Denny Zhou, Nathanael Scharli, Le~Hou, Jason Wei, Nathan Scales, Xuezhi Wang,
  Dale Schuurmans, Olivier Bousquet, Quoc Le, and Ed~Huai hsin Chi. 2023.
\newblock {Least-to-Most Prompting Enables Complex Reasoning in Large Language
  Models}.
\newblock In \emph{Proceedings of ICLR}.

\end{thebibliography}
